\documentclass[10pt,twocolumn,letterpaper]{article}

\usepackage{cvpr}
\usepackage{times}
\usepackage[dvipdfmx]{graphicx}
\usepackage{amsmath}
\usepackage{amssymb}

\usepackage{amsfonts}
\usepackage{color}
\usepackage{comment}

\usepackage[pagebackref=true,breaklinks=true,letterpaper=true,colorlinks,bookmarks=false]{hyperref}

\cvprfinalcopy 

\newcommand{\citet}{\cite}
\newcommand{\citep}{\cite}

\newcommand{\putFigW}[3]{%
  \begin{figure}[]%
    \centering%
    \includegraphics[width=#3]{fig/#1.pdf}%
    \caption{#2}%
    \label{fig:#1}%
  \end{figure}}

\newcommand{\putFigWW}[3]{%
  \begin{figure*}[]%
    \centering%
    \includegraphics[width=#3]{fig/#1.pdf}%
    \caption{#2}%
    \label{fig:#1}%
  \end{figure*}}
  
\newcommand{\putFigWWH}[3]{%
  \begin{figure*}[]%
    \centering%
    \includegraphics[width=#3]{fig/#1.pdf}%
    \caption{#2}%
    \label{fig:#1}%
  \end{figure*}}

\newcommand{\refFig}[1]{{Fig. \ref{fig:#1}}}
\newcommand{\refEq}[1]{Eq. (\ref{eq:#1})}

\ifcvprfinal\fi
\begin{document}

\title{Spherical Image Generation from a Single Normal Field of View Image by Considering Scene Symmetry}

\author{Takayuki Hara$^1$ and Tatsuya Harada$^{1,2}$\\
$^1$The University of Tokyo, $^2$Riken\\
{\tt\small \{hara, harada\}@mi.t.u-tokyo.ac.jp}
}

\maketitle

\begin{abstract}
Spherical images taken in all directions (360$^\circ \times $180$^\circ$) allow representing the surroundings of the subject and the space itself, providing an immersive experience to the viewers. Generating a spherical image from a single normal-field-of-view (NFOV) image is convenient and considerably expands the usage scenarios because there is no need to use a specific panoramic camera or take images from multiple directions; however, it is still a challenging and unsolved problem. The primary challenge is controlling the high degree of freedom involved in generating a wide area that includes the all directions of the desired plausible spherical image. On the other hand, scene symmetry is a basic property of the global structure of the spherical images, such as rotation symmetry, plane symmetry and asymmetry. We propose a method to generate spherical image from a single NFOV image, and control the degree of freedom of the generated regions using scene symmetry. We incorporate scene-symmetry parameters as latent variables into conditional variational autoencoders, following which we learn the conditional probability of spherical images for NFOV images and scene symmetry. Furthermore, the probability density functions are represented using neural networks, and scene symmetry is implemented using both circular shift and flip of the hidden variables. Our experiments show that the proposed method can generate various plausible spherical images, controlled from symmetric to asymmetric.
\end{abstract}

\section{Introduction}

\putFigW{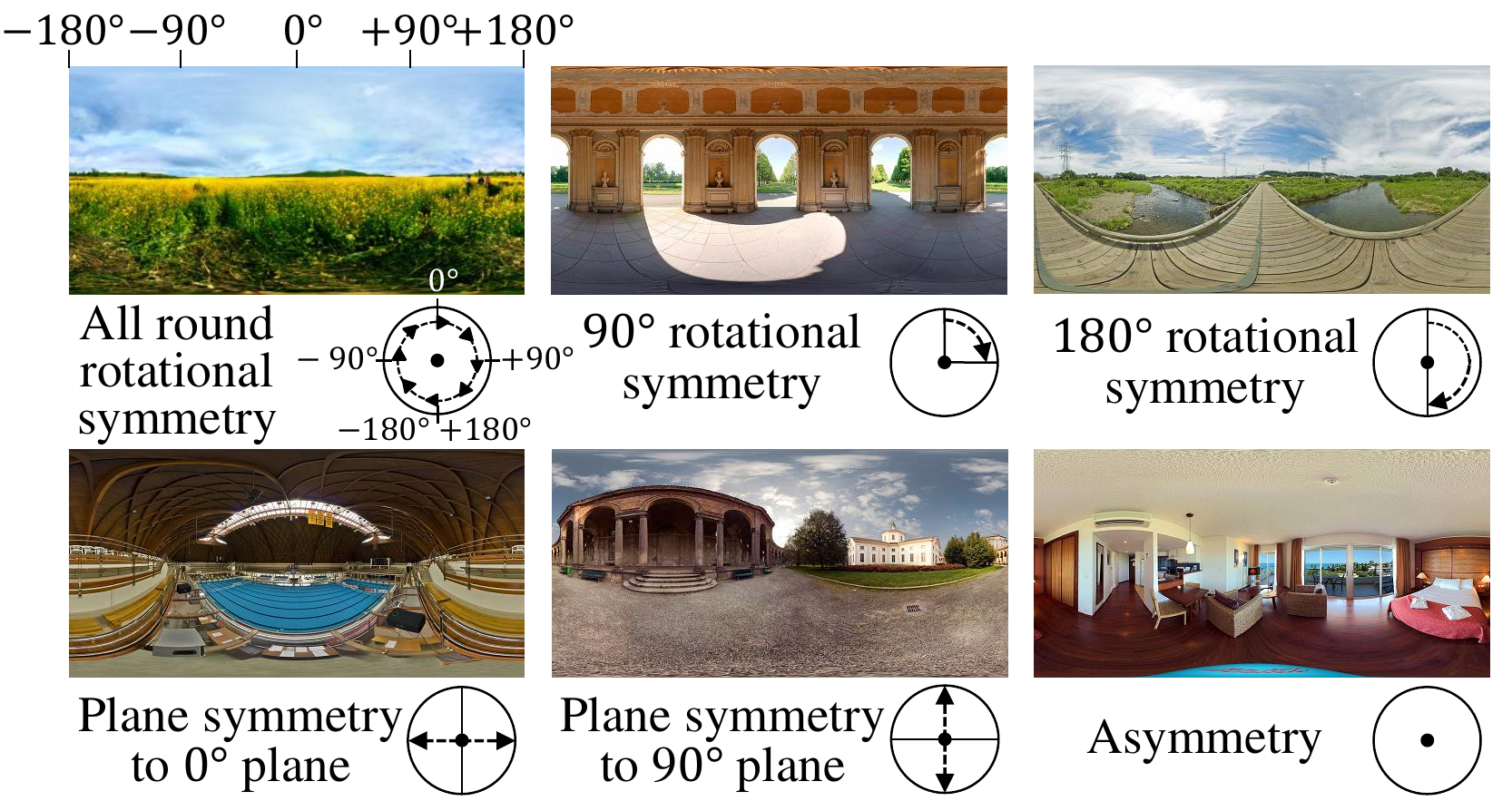}{Symmetry types of spherical images in SUN360 dataset \cite{viewpoint}. Images are represented by equirectangular projection (see \refFig{equi}), and arrows in the circles correspond to the viewpoint transitions that do not change the appearance significantly.}{85mm}

Spherical images, which capture all possible directions (horizontal 360$^\circ$ $\times$ vertical 180$^\circ$), are used in various domains such as surveillance systems, building industry, tourism, autonomous cars, and entertainment. They can capture environments around the main subject and represent the space itself as a subject. Furthermore, when viewed through a head-mounted display, spherical images allow one to enjoy a scene in a more immersive manner. However, capturing spherical images is not an easy task, as doing so requires a specific panoramic camera or specific software that stitches together images taken from multiple directions. Therefore, it would be more convenient to generate a spherical image from a single normal-field-of-view (NFOV) image taken using a normal camera. Furthermore, it would considerably expand the usage scenarios that require plausibility rather than reproducibility; for example, the background of VR content can be created using a single concept photo, or generating peripheral views of pictures taken in the past or historic paintings can allow the viewers to enjoy the content with immersive feeling.

However, two main challenges exist while generating a spherical image from a single NFOV image: one is generating an image corresponding to the spherical structure, and the other is controlling the high degree of freedom involved in generating a wide area, which includes the all directions of a plausible spherical image. Because a spherical image cannot be uniquely determined for an NFOV image, the generation of various plausible scenes must be controlled. 

Generating a spherical image from an NFOV image is related to the task of image completion, which predicts a whole image from a partial one. However, conventional image-completion methods \cite{GAN_IC1,GAN_IC2,PIC} are not suitable for generating spherical images, because these methods are designed for planar NFOV images. Recently, a spherical-image-completion method using a single NFOV image was proposed \cite{360gen}. This method can handle the spherical structure by performing rearrangement in an equirectangular projection for the input and employing dilated convolution; however, it cannot control the content of the generated regions. Therefore, unlike conventional research works, we handle the spherical structure and aim to control the aforementioned degree of freedom in order to obtain the plausible variations of the desired spherical image.

Various factors control spherical-image generation. Among these factors, {\it scene symmetry} is a basic property of the global structure of spherical images. As pointed out in \cite{viewpoint}, the 360$^\circ$ view has specific symmetry types, meaning that specific geometric operations such as rotation and flip significantly change the appearance or not. \refFig{symmetry} depicts the typical types of symmetry,  such as rotational symmetry, plane symmetry and asymmetry. 
When a spherical image possesses a certain level of symmetry, a correlation exists between specific areas of the image. On the basis of this observation, we control the scene-symmetry properties to improve the plausibility and make diverse generation. Furthermore, we control the symmetry intensity, as a perfectly symmetric image is sometimes unnatural. 

The contributions of this paper can be summarized as follows.
\begin{itemize}
\item We propose a novel spherical-image-generation method using a single NFOV image; this method improves the plausibility of the generated images by leveraging scene symmetry.
\item We design a new architecture that is able to estimate and control the symmetry of the image by using the circular shift and the flip of the hidden variables of convolutional neural networks (CNNs).
\item We demonstrate that our proposed method can generate multiple spherical images, controlled from symmetric to asymmetric.
\end{itemize}

\section{Related Work}
\subsection{Image Completion}
Various image-completion technologies have been proposed thus far in order to predict the missing regions of an image. Traditionally, several diffusion-based methods \cite{diffusion1,diffusion2} diffuse the information of the visible regions to the missing regions, and several patch-based methods \cite{patch1,patch2} complete the missing regions by matching, copying, and realigning using the visible regions. Both these methods assume that the missing regions contain information correlated with the visible regions.

Recently, generative models, which are learned using large-scale datasets, such as the variational autoencoder (VAE) \cite{VAE,VAE2} and generative adversarial networks (GANs) \cite{GAN} have experienced a significant boost, and both of these models have been adopted for image completion. Li {\it et al.} \citet{GAN_IC1} directly generated the contents for missing regions on the basis of CNNs \cite{CNN1,CNN2} with a combination of a reconstruction loss, a semantic parsing loss, and two adversarial losses. Iizuka {\it et al.} Furthermore, \citet{GAN_IC2} employed global and local context discriminators in the framework of adversarial learning to improve the naturalness and consistency of the completed regions.

While most image-completion methods produce only one result for each input, Zheng {\it et al.} \citet{PIC} presented a method to generate multiple and diverse plausible solutions for image completion by using conditional VAEs (CVAEs) \cite{CVAE}; however, their method cannot explicitly control the generated contents. Therefore, unlike any of the previous methods, we propose a spherical-image-generation method that can explicitly control the symmetry.

\subsection{Panoramic and Spherical Image Generation}
Zhang {\it et al.} \citet{framebreak} proposed to extrapolate an NFOV image to a panoramic image, with the guidance of a panorama image. The method required a guide image belonging to the same scene category to which the input image belonged. Kimura {\it et al.} \citet{ExtVision} proposed a peripheral-image-generation method based on pix2pix \cite{pix2pix}; however, the field of view of the generated image was limited. Sumantri {\it et al.} \citet{panoSyn} proposed a spherical-image-generation method based on pix2pixHD \cite{pix2pixhd}, which required a set of images captured from multiple directions as an input.  Recently, Akimoto {\it et al.} \cite{360gen} proposed  a spherical-image-completion method using a single NFOV image. This method can handle the spherical structure by means of rearrangement in an equirectangular projection for the input and employing dilated convolution; however, it cannot control the content of the generated regions. Unlike these conventional research works, we handle the spherical structure and aim to control the aforementioned degree of freedom to obtain plausible variations of the desired spherical image.

\putFigW{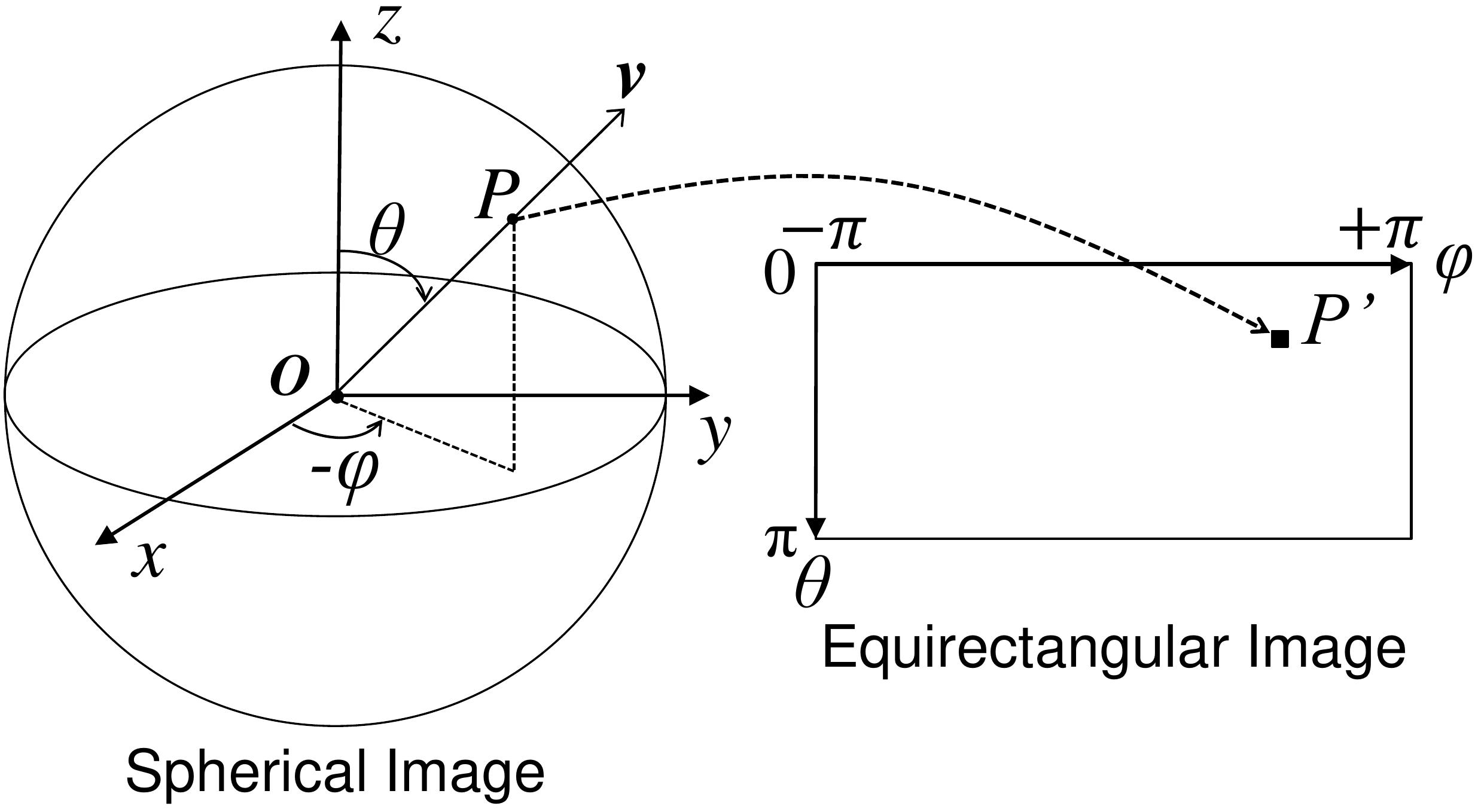}{Equirectangular projection. Point $P$ on the 2D sphere is converted to point $P '$ on the $\theta \mathchar`- \phi$ plane.}{70mm}

\subsection{CNN for Spherical Signals}
Ordinal 2D CNNs are designed on the basis of translational equivariance on a plane; therefore they are not directly applied to spherical images. A CNN-based approach for generating spherical signals is to apply planar CNNs in the projected plane, such as an equirectangular projection \cite{equiCNN,saliency360} depicted in \refFig{equi}, tangent planes \cite{tateno}, and cube mapping \cite{cubeCNN}. Other approaches that do not use 2D planar CNNs, such as generalized FFT-based CNNs on sphere \cite{SCNN1,SCNN2}, distortion-aware sampling \cite{DAS1,DAS2}, and graph convolution on sampled grid on sphere \cite{GCNN1,GCNN2}, were also proposed. 

Although projection distortion is caused when using the methods based on planar CNNs, some good spherical images were generated by using the equirectangular projection in \citet{panoSyn,360gen}. Furthermore, using the equirectangular projection offer advantages such as the symmetry operation around the gravity axis can be realized using the circular shift and the flip with low computational requirements. In this study, we employ equirectangular-projection-based CNNs.

\section{Proposed Method}
This section describes our proposed method, which generates a spherical image $x_g$ from a partial image (e.g., an NFOV image) $x_l$ by using a scene-symmetry parameter $s \in \mathbb{R}^C$. The term $s$ corresponds to the intensity of $C$ types of rotational symmetry  around the gravity-axis, and plane symmetry to vertical planes (the gravity-axis is aligned with the negative direction of the z-axis in \refFig{equi}). To generate diverse images conditioned with the partial image, we employ CVAEs \cite{CVAE} as a base framework, and we also incorporated scene-symmetry parameters as latent variables. Furthermore, the probability density functions are represented by the equirectangular projection-based CNNs, and scene symmetry is implemented as a circular shift and a flip of the hidden variables. The following describes the details of our method.

\subsection{CVAEs with Symmetry Parameters}
First, we describe our probabilistic framework.  To generate diverse spherical images conditioned with the partial image, we employ CVAEs \cite{CVAE} as a base framework. We obtain the conditional distribution $p(x_g | x_l, s)$, maximizing the variational lower bound of the likelihood for the training data $\{x_g^{(n)}, x_l^{(n)}\}_{n=1}^N$ ($s$ is not required), following which we sample a new $x_g$ for the given $x_l$ and $s$ from this distribution.
 
In CVAEs, the variational lower bound of the conditional log-likelihood of observing a partial image $x_l$ for a spherical image $x_g$ is expressed as follows:
\begin{eqnarray}
\label{eq:cvae}
\log p(x_g | x_l) \geq  -{\rm KL}(q_\psi(z'|x_g, x_l) || p_\phi(z'|x_l)) \nonumber \\
  + {\rm E}_{q_\psi(z'|x_g, x_l)} [\log p_\omega(x_g | x_l, z')]
\end{eqnarray}
where $z'$ denotes the latent vector with the whole image information encoded; $q_\psi(\cdot|\cdot)$ the posterior importance sampling function; $p_\phi(\cdot|\cdot)$ the conditional prior; $p_\omega(\cdot|\cdot)$ the likelihood; $\psi$, $\phi$, and $\omega$ the parameters of the neural networks; ${\rm KL}(q||p)$ the KL divergence between $q$ and $p$; and ${\rm E}_q[\cdot]$ the expected value for the probability distribution $q$. Each probability distribution is obtained by determining the parameters so as to maximize the variational lower bound.

Furthermore, we introduce the scene-symmetry parameter $s$ as a part of the latent vector $z_g$, i.e., $z' = (z, s)$, and we let $s$ be independent of the partial image $x_l$, i.e., $ p_\phi(z, s|x_l) =  p_\phi(z|x_l)p(s)$. Thus, \refEq{cvae} can be written as follows:
\begin{eqnarray}
\label{eq:cvaes}
\log p(x_g | x_l) \geq  -{\rm KL}(q_\psi(z, s|x_g, x_l) || p_\phi(z|x_l)p(s)) \nonumber \\
  + {\rm E}_{q_\psi(z, s|x_g, x_l)} [\log p_\omega(x_g | x_l, z, s)].
\end{eqnarray}
The parameters $\psi$, $\phi$, and $\omega$ are determined to maximize the variational lower bound for the training data $\{x_g^{(n)}, x_l^{(n)}\}_{n=1}^N$. We obtain the targeted conditional distribution as follows:
\begin{equation}
\label{eq:samp}
p(x_g | x_l, s) = \int p_\omega(x_g | x_l, z, s) p_\phi(z|x_l) dz
\end{equation}
where $x_g$ is sampled from this distribution for the given $x_l$ and $s$.

\subsection{Likelihood for Reconstruction and Generation}
We assign two kinds of functions for the log-likelihood $\log p_\omega(x_g | x_l, z, s)$. The first function is the reconstruction likelihood $l_\omega^{\rm rec}(x_g, x_l, z, s)$ to reconstruct $x_g$ corresponding to $x_l$ in the training data. The second function is the generation likelihood $l_\omega^{\rm gen}(x_l, z, s)$ to generate a sample that does not depend on specific $x_g$ but follows the distribution of the training data (we allow these likelihood functions deviate from the probability distribution). Thereafter, we maximize the combination of these two functions to generate the samples that possess both plausibility and diversity. The combination of the two criteria, namely, the reconstruction and generation errors, is inspired by PICNet \cite{PIC}. However, our method differs with respect to the following important point: the scene-symmetry parameter is incorporated to latent variables.

First, to maximize the reconstruction likelihood, we employ the following decomposition approximation, because it is difficult to perform the maximization directly. We have
\begin{equation}
\label{eq:app1}
q_\psi(z, s|x_g, x_l) = q_{\psi}(z|x_g) q_{\psi}(s|x_g)
\end{equation}
Using the approximation in \refEq{app1}, the right-hand side of \refEq{cvaes} can be transformed as follows:
\begin{eqnarray}
\label{eq:app1b}
{\mathcal L}_{\rm rec}(x_g | x_l) \!\!\!\!&=&\!\!\!\! -{\rm KL}(q_{\psi}(z |x_g) || p_\phi(z|x_l)) \nonumber \\
&&\!\!\!\!-{\rm KL}(q_{\psi}(s |x_g) || p(s))  \nonumber \\
&&\!\!\!\!+ {\rm E}_{q_{\psi}(z|x_g) q_{\psi}(s|x_g)} [l_\omega^{\rm rec}(x_g, x_l, z, s)]
\end{eqnarray}
However, for the generation likelihood, we  approximate $q_\psi(z, s|x_g, x_l)$ by removing the dependency to $x_g$ as follows:
\begin{equation}
\label{eq:app2}
q_\psi(z, s|x_g, x_l) = q_{\psi}(z|x_l) q_{\psi}(s).
\end{equation}
Using the approximation in \refEq{app2}, as ${\rm KL}(q || p )$ takes the minimum value of 0 when $q = p$, the right-hand side of \refEq{cvaes} can be transformed as follows:
\begin{eqnarray}
\label{eq:app2b}
{\mathcal L}_{\rm gen}(x_g | x_l) = {\rm E}_{p_\phi(z|x_l)p(s)} [l_{\omega}^{\rm gen}(x_l, z, s)]
\end{eqnarray}
We combine both these likelihoods and maximize the following evaluation function by adding each variational lower bound with the ratio of $ \gamma \in [0, 1]$. We have
\begin{equation}
\label{eq:eval}
{\mathcal L}(x_g | x_l) =  \gamma {\mathcal L}_{\rm rec}(x_g | x_l) + (1 - \gamma) {\mathcal L}_{\rm gen}(x_g | x_l)
\end{equation}
We determine the parameters $\psi, \phi$ and $\omega$ to maximize the summation of ${\mathcal L}(x_g | x_l)$ for the training data $\{x_g^{(n)}, x_l^{(n)}\}_{n=1}^N$. Details of the probabilistic framework are described in the Appendix \ref{sec:prob_framework}.

\subsection{Probability Distribution Settings}
The prior and posterior distributions for the latent variables $z$ and $s$ are set as follows:
\begin{eqnarray}
\label{eq:q_z}
q_{\psi}(z |x_g) \!\!\!\!&=&\!\!\!\! {\mathcal N}(z | F_\mu \circ F(x_g), F_{\Sigma} \circ F(x_g))\\
\label{eq:p_z}
p_\phi(z|x_l) \!\!\!\!&=&\!\!\!\!  {\mathcal N}(z | \tilde{F}_\mu \circ F(x_l),\tilde{F}_{\Sigma} \circ F(x_l))\\
\label{eq:p_s}
q_{\psi}(s |x_g) \!\!\!\!&=&\!\!\! \delta(s - F_s \circ F(x_g ))\\
\label{eq:p_s}
p(s) \!\!\!\!&=&\!\!\!\!  {\mathcal N}(s | \mu_s, \Sigma_s)
\end{eqnarray}
where $ {\mathcal N}(\cdot | \mu, \Sigma)$ denotes the probability density function of a normal distribution with the mean vector $\mu$ and the covariance matrix $\Sigma$; $\delta(\cdot)$ denotes a delta function;  and $\mu_s$ and $\Sigma_s$ denote the hyperparameters of the mean vector and the covariance matrix of $s$, respectively. Furthermore, $F$ denotes the function for image-feature extraction, and $F_\mu, F_{\Sigma}, \tilde{F}_\mu, \tilde{F}_{\Sigma}, F_s$ denote the functions that estimate the parameters of each distribution. In our implementation, $F_{\Sigma}, \tilde{F}_{\Sigma}$ output the diagonal matrices, indicating that each element of $z$ is conditionally independent in $q_{\psi_z}(z |x_g)$ and $p_\phi(z|x_l)$.

In addition, the likelihood functions are set as follows:
\begin{eqnarray}
\label{eq:p_x}
l_{\omega}^{\rm rec}(x_g, x_l, z, s) \!\!\!\!&=&\!\!\!\!  \alpha||D(x_g) - D(G(x_l, z, s))||_2 \nonumber \\
&&\!\!\!\! + \beta||x_g - G(x_l, z, s)||_1 \\
\label{eq:p_xx}
l_{\omega}^{\rm gen}(x_l, z, s) \!\!\!\!&=&\!\!\!\! \ \alpha||\mathbf{1} - D(G(x_l, z, s))||_2^2 \nonumber \\
&&\!\!\!\! + \beta||x_l - M \circ G(x_l, z, s))||_1
\end{eqnarray}
where $\alpha, \beta (\leq 0)$ denote the weighting factors, $G$ is a function that generates a spherical image, $D$ is a function that outputs confidences to discriminate real images for multiple partial regions, $\mathbf{1}$ is a constant vector whose elements take all the value of 1, and $M$ is a function that extracts the image in the region corresponding to $x_l$.
\subsection{Network Structure}

\putFigW{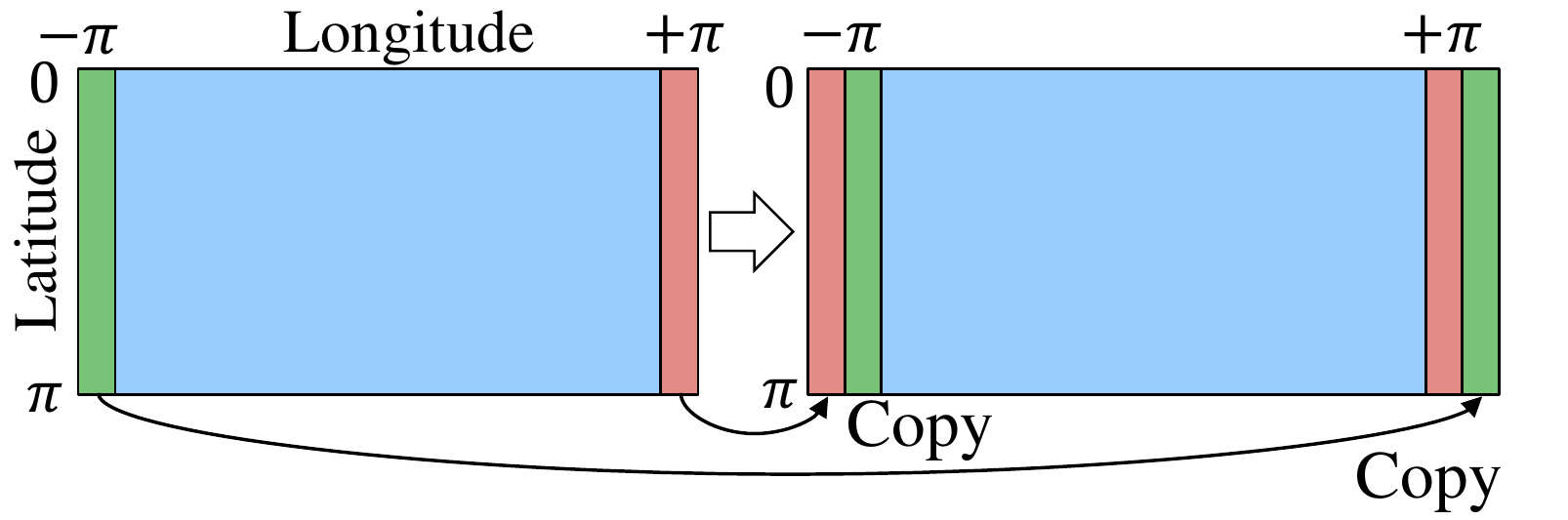}{Circular padding. The circular padding eliminates the discontinuity between the left and right edges of the equirectangular image for convolution.}{62mm}

\putFigWW{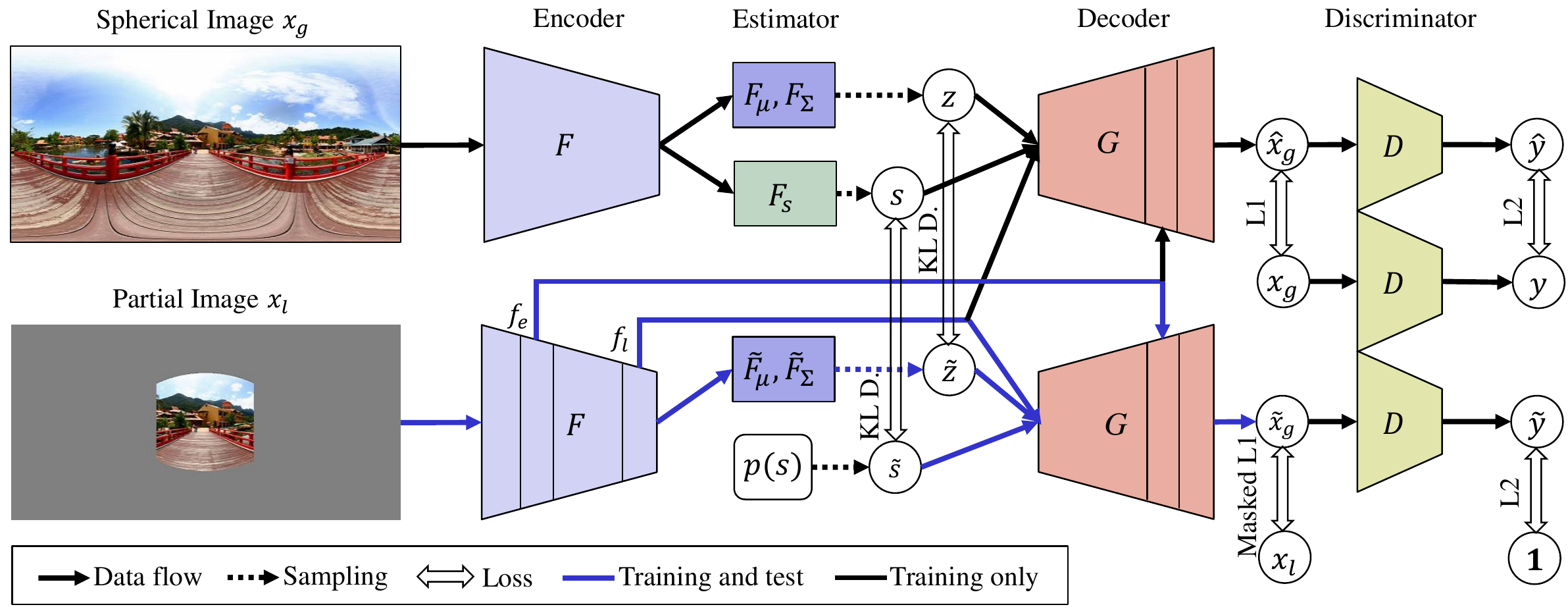}{Structure of our proposed method for spherical-image generation. During training, the spherical image $x_g$ and the partial (NFOV) image $x_l$ are used. During testing, a spherical image $x'_g$ is generated from a single $x_l$.}{170mm}

Functions $F, F_\mu, F_\Sigma, F_s, \tilde{F}_\mu, \tilde{F}_\Sigma, G, D$ are implemented using fully CNNs. To eliminate the discontinuity between the left and right edges of the equirectangular image, we employ the circular padding before each convolution layer. As depicted in \refFig{c_padding}, the circular padding copies $(k-1)$ $\rm{mod}$ $2$ columns from the left and right edges of the image to each opposite side in the equirectangular image, where $k$ denotes the kernel size of the convolution. The implementation detail of each function is described in Section \ref{sec:detail}. 

\refFig{net_all} depicts the entire network structure of the proposed method. The input spherical image $x_g$ and the partial image $x_l$ are represented using equirectangular projections (see \refFig{equi}). The encoder $F$ and the estimators $F_\mu, F_{\Sigma}, F_s$ calculate the latent variables $z$ and $s$ for $x_g$; in addition, $F, \tilde{F}_\mu$ and $\tilde{F}_{\Sigma}$ calculate $\tilde{z}$ for $x_l$ using the {\it reparameterization trick} \cite{VAE}. Moreover, $F$ outputs the partial features $f_e, f_l$ for $x_l$, where $f_e$ and $f_l$ are obtained from the hidden layer and the last layer, respectively. In addition, $\tilde{s}$ is sampled from $p(s)$. The decoder $G$ generates the spherical images $\hat{x}_g$ from $f_e, f_l, z$ and $s$, and $\tilde{x}_g$ from $f_e, f_l, \tilde{z}$ and $\tilde{s}$. The scene symmetry $s$ is estimated on the basis of the rotated and flipped differences of the hidden variables of $F_s$, and the hidden variables of $G$ are circular shifted, flipped, and copied depending on $s$, as described in Section \ref{sec:symmetry}. In this manner, the symmetry of the spherical image is encoded and decoded. The discriminators $D$ output the confidences $\hat{y}$ and $\tilde{y}$ from $\hat{x}_g$ and $\tilde{x}_g$ in order to discriminate real images. Thereafter, the loss is calculated according to \refEq{eval}. 

\subsection{Estimating and Controlling Scene Symmetry}
\label{sec:symmetry}

\putFigW{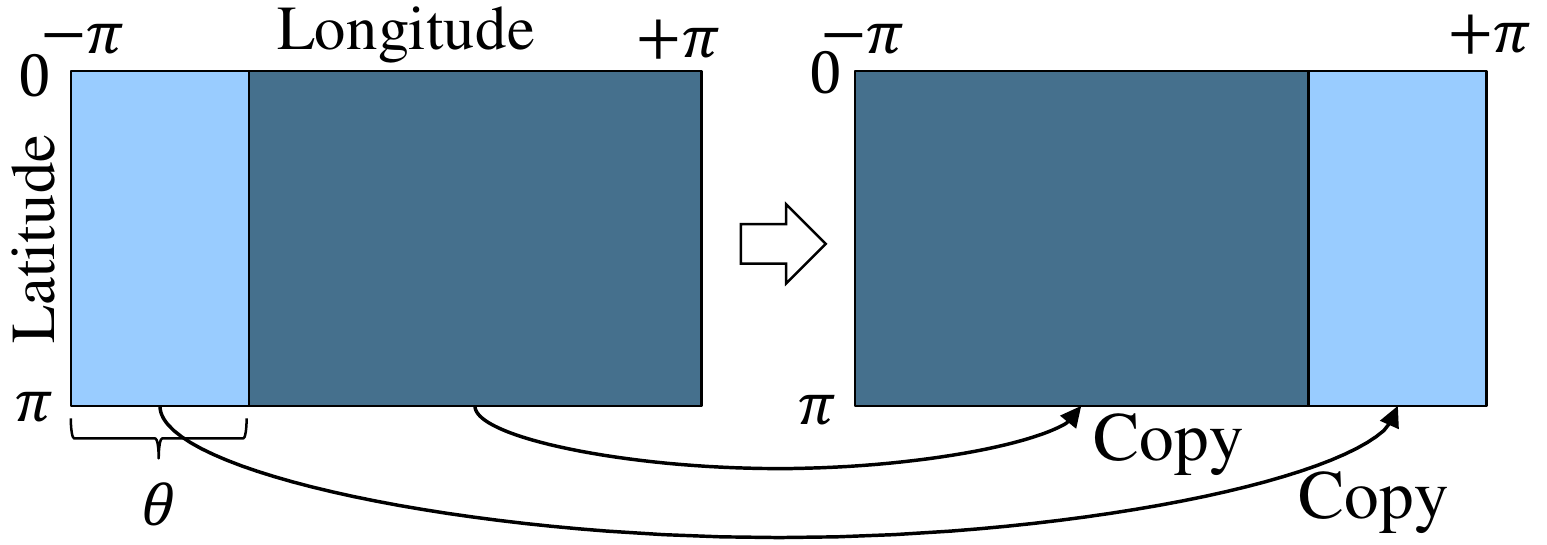}{Circular shift. The circular shift is used for estimating and controlling the rotational symmetry.}{62mm}

\putFigW{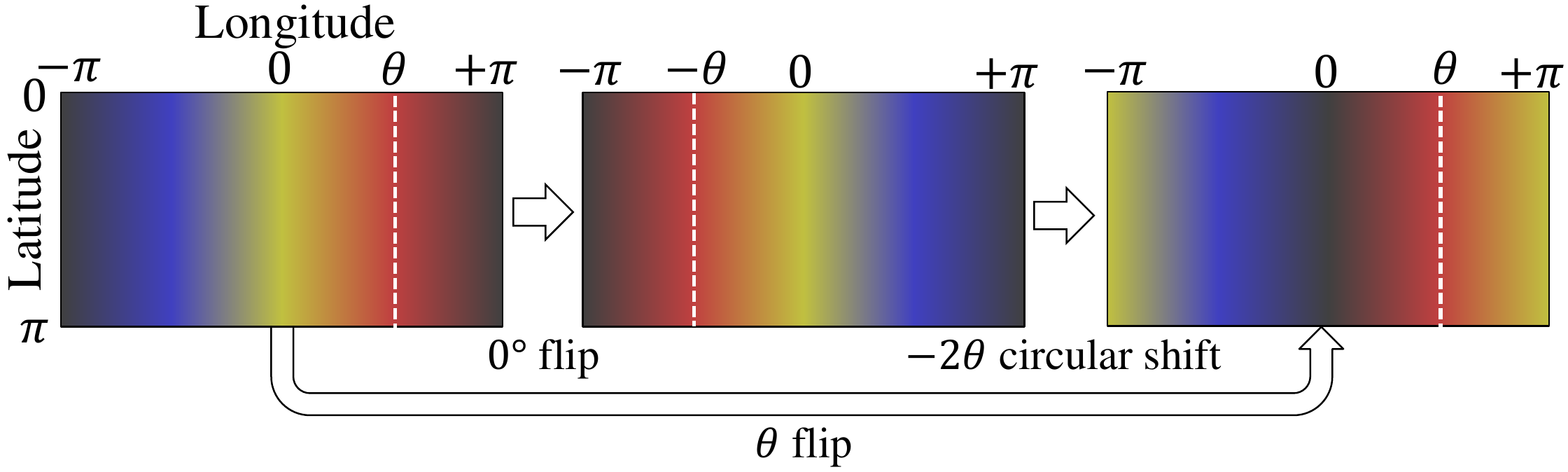}{Flip on the $\theta$ axis. This operation combines 0$^\circ$ flip and $-2\theta$ circular shift. The flip is used for estimating and controlling the plane symmetry.}{84mm}

We propose a novel method for estimating and controlling the scene symmetry for spherical images by using both a circular shift and a flip in the equirectangular projection. Because our networks consist of fully CNNs, the hidden variables store the location information of the input equirectangular image. In the equirectangular projection, a rotation about the gravity-axis corresponds to a horizontal circular shift, as depicted in \refFig{c_shift}, and a flip to the longitude $\theta$ plane corresponds to 0$^\circ$ flip and $-2\theta$ circular shift, as depicted in \refFig{flip}. In this manner, we represent both rotational and plane symmetries via the circular shift and the flip of the hidden variables, respectively.

The estimator $F_s$ calculates the hidden variables $f_s$, following which $s = (s_1, s_2, \cdots, s_C)$ is calculated formulas follows:
\begin{equation}
\label{eq:estimate_s}
s_i = \sigma(\zeta ||T_i (f_s) - f_s||_1 + \eta)
\end{equation}
where $\sigma(\cdot)$ denotes a sigmoid function, $T_i$ a symmetric-transformation function consisting of the circular shift and flip, and $\zeta$ and $\eta$ the trainable parameters.

The decoder G controls the scene symmetry of the generated image. First, we define the symmetry-control function $H$, which takes the weighted linear sum of the symmetric transformation of the input feature $f$ as follows:
\begin{equation}
\label{eq:control}
H(f, s) = \frac{w_{\kappa} \odot f + \sum_{i=1}^C s_i T_i (w_{\kappa} \odot f)}{w_{\kappa} +  \sum_{i=1}^C s_i T_i (w_{\kappa})}
\end{equation}
where $w_{\kappa}$ denotes a weight vector, $\odot$ an elementwise product, and the quotient is elementwise. In addition, the element of $w_{\kappa}$ corresponding to the position $v$ on the sphere is defined as follows:
\begin{equation}
\label{eq:weight}
w_{\kappa}(v) = \exp (\kappa \langle v, c \rangle)
\end{equation}
where $\kappa$ denotes a hyper parameter for the concentration, $c$ the center position of the input partial image on the sphere, and $\langle \cdot, \cdot \rangle$ the inner product of the 3D Euclidean space. That is, a high weight is applied at the position of the partial image and attenuates as it goes around. 

The scene symmetry is reflected in the generated image as follows. The decoder $G$ takes $H(f_l + z, s)$ as the input of the first layer and, subsequently, concatenates $H(f_e, s)$ to the output of the specified hidden layer in the channel dimension. Thereafter, $G$ outputs the spherical image $x_g$.

\subsection{Adversarial Learning}
We employ adversarial learning combined with VAE, as used in \cite{VAE-GAN,CVAE-GAN,PIC}. First, we learn $F, F_{\mu}, F_{\Sigma}, F_{s}, \tilde{F}_{\mu}, \tilde{F}_{\Sigma}, G$ to maximize the evaluation function of \refEq{eval}. Subsequently, on the basis of LSGAN \cite{LSGAN} for each mini-batch data, we learn $D$ to minimize the following loss function:
\begin{equation}
\label{eq:l_d}
{\mathcal L}_D = {\rm E}_{p(x_g)}[||\mathbf{1} - D(x_g) ||_2^2]  +  {\rm E}_{p(x_g|x_l, s)}[||D(x_g)||_2^2] 
\end{equation}

\subsection{Sampling-Generated Spherical Image}
After training, we obtain $p(x_g | x_l, s)$ in Eq. (\ref{eq:samp}), following which we sample $x_g$ from this distribution for the given $x_l$ and $s$ by using the following two steps: (i) sampling $z$ from $p_\phi(z|x_l)$ in \refEq{p_z}, and (ii) sampling $x_g$ from $p_\omega(x_g | x_l, z, s)$ corresponding to $l_\omega^{\rm rec}$ in \refEq{p_x}. Because $l_\omega^{\rm rec}$ does not represent probability distribution, instead, we take a sample to maximize $l_\omega^{\rm rec}$, i.e., $G(x_l, z, s)$.

\putFigWWH{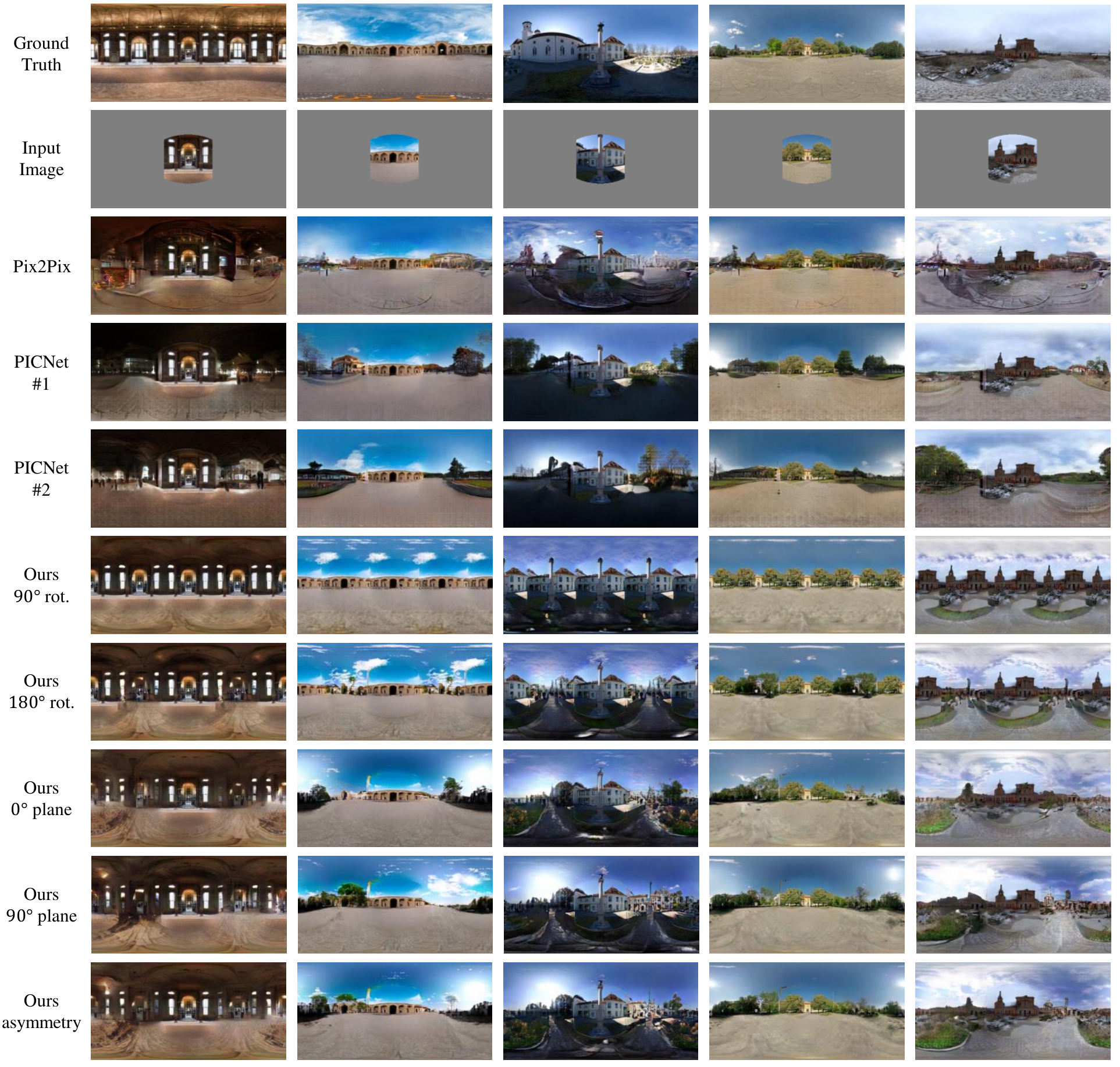}{Qualitative comparison between spherical images generated using our method and those generated using previous works. Our method enables scene-symmetry control.}{175mm}

\section{Experiments}
We conducted experiments to verify the effectiveness of the proposed method. To that end, we used the Sun360 dataset \cite{viewpoint}, which includes various spherical images, both symmetric and asymmetric, from indoor to outdoor. The data were divided into 50,000 images for training, 10,000 images for testing, and 5,000 images for validation. The spherical image was an RGB image of the equirectangular format with a resolution of $256 \times 512$ pixel. Furthermore, a partial (NFOV) image was cropped from the spherical image with 30$^\circ$ to 120$^\circ$ field of view and aspect ratio of 1: 1, following which the viewpoint direction was set randomly on the sphere and projected to the equirectangular image, whose margin was filled with gray values.

\subsection{Implementation Details}
\label{sec:detail}

\putFigW{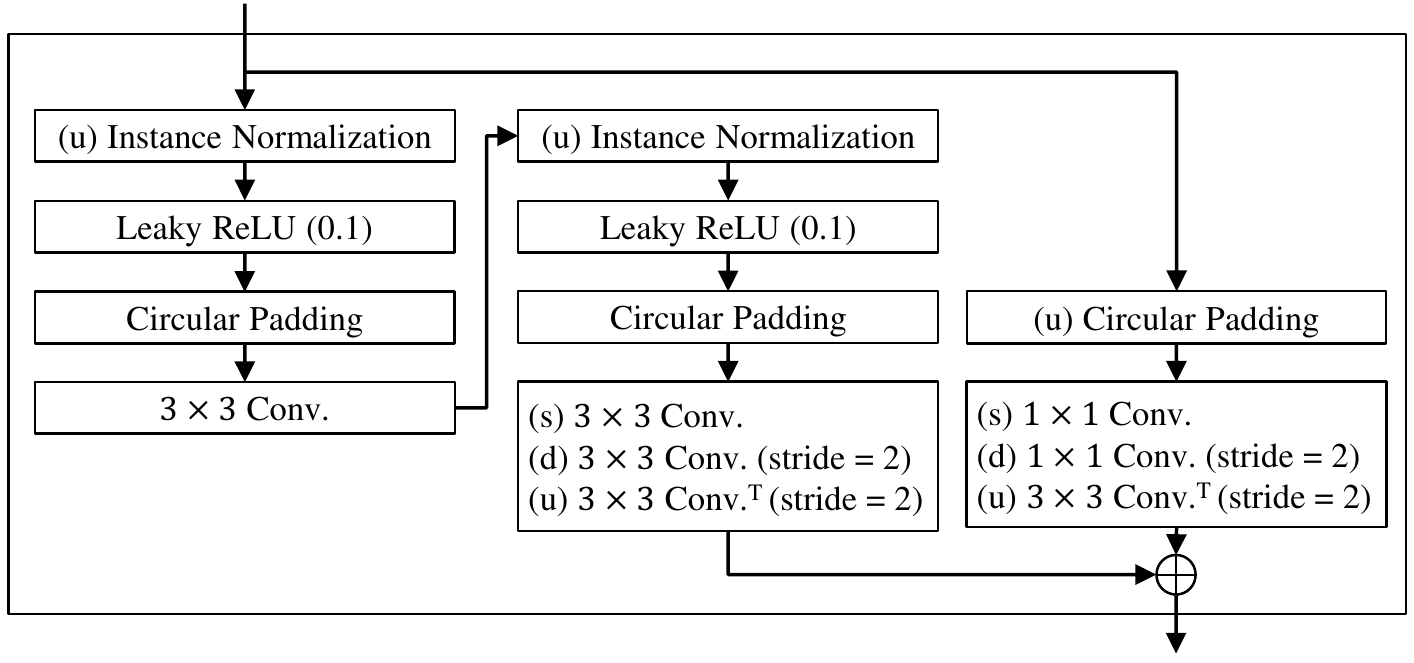}{Residual block with circular padding. There are three modes: (s) standard, (d) down-sampling, and (u) up-sampling. For example, during the standard mode, only the modules marked with (s) are used.}{84mm}

We trained the networks from scratch using Adam optimizer \cite{adam} with a fixed learning rate of $10^{-4}$ and a mini-batch size of 8. During the optimization, the weighting factors of the likelihood were set to $\alpha=-1.0$ and $\beta=-20.0$; the priors for symmetry were set to $\mu_s = 0.50$ and $\sigma_s = 0.33$; the concentration parameter $\kappa$ was set to 3.0; the mixture ratio of the two approximations was set to $\gamma=0.5$. Furthermore, we considered $C=5$ types of symmetry, namely the 90$^\circ$, 180$^\circ$ and 270$^\circ$ rotational symmetries and plane symmetries to the 0$^\circ$ and 90$^\circ$ axes.

Our network consisted of a multi-layer ResBlock with circular padding (RBCP), which is a unit module of residual networks \cite{ResNet} with circular padding, as depicted in \refFig{srb}. The RBCP has three modes: (s) standard, (d) down-sampling, and (u) up-sampling, and the behavior of each of these is different, as depicted in \refFig{srb}.
The structure of each function is as follows:

$\bullet$ $F$: two-layer RBCP(s) and five-layer RBCP(d).

$\bullet$ $F_\mu$ and $F_\Sigma$: one-layer RBCP(s).

$\bullet$ $\tilde{F}_\mu$ and $\tilde{F}_\Sigma$: seven-layer RBCP(s) and share the weights of the first six layers. 

$\bullet$ $F_s$: one-layer RBCP(d).

$\bullet$ $G$: one-layer RBCP(s) and five-layer RBCP(u).

$\bullet$ $D$: five-layer RBCP(d) and $3 \times 3$ conv. with 1 output channel.

The encoder $F$ outputs $f_e$ and $f_l$, which are outputs of the fifth and final (seventh) layers, respectively. Furthermore, we set the size (height, width, and channel) of $f_e, f_l, z$, and $f_s$ to $32 \times 64 \times 128$, $8 \times 16 \times 128$, $8 \times 16 \times 128$, and $4 \times 8 \times 128$, respectively. After the third layer of $G$ and $D$, we employed {\it self-attention} \cite{SAGAN} to harness the distant spatial context. Both $D$ output the confidences for $6 \times 6$ partial regions. These configurations were matched to PICNet \cite{PIC} for a comparison.

\subsection{Comparison with Related Works}
We compare our method with pix2pix \cite{pix2pix} and PICNet \cite{PIC}. Pix2pix has been used for peripheral-image generation \cite{ExtVision} and as a basis for spherical-image generation \cite{360gen}. PICNet is a state-of-the-art image-completion method; its input comprises partial equirectangular images, similar to the proposed method. Additionally, circular padding is added before each convolution layer.

\subsubsection{Qualitative Evaluation}
\refFig{sample} depicts the examples of spherical images generated from the frontal 90$^\circ$ views by using each method. In our proposed method, the symmetry parameter $s \in \mathbb{R}^5$ was set to five variations, namely, $(h, h, h, l, l)$, $(l, h, l, l, l)$, $(l, l, l, h, l)$, $(l, l, l, l, h)$ and $(l, l, l, l, l)$ where $h=1.0, l=0.3$, for the multiple of 90$^\circ$ and 180$^\circ$ rotational symmetries, the plane symmetries to the 0$^\circ$ and 90$^\circ$ axes, and asymmetry, respectively. Furthermore, PICNet generates multiple samples using randomized latent variables. Although PICNet can generate a variety of images, it cannot explicitly control the generated contents. For example, unlike our method, none of the generated images using PICNet possesses a spatial structure that significantly differs from that of the rest. More generation results are available in the Appendix \ref{sec:additional_examples}.

\putFigW{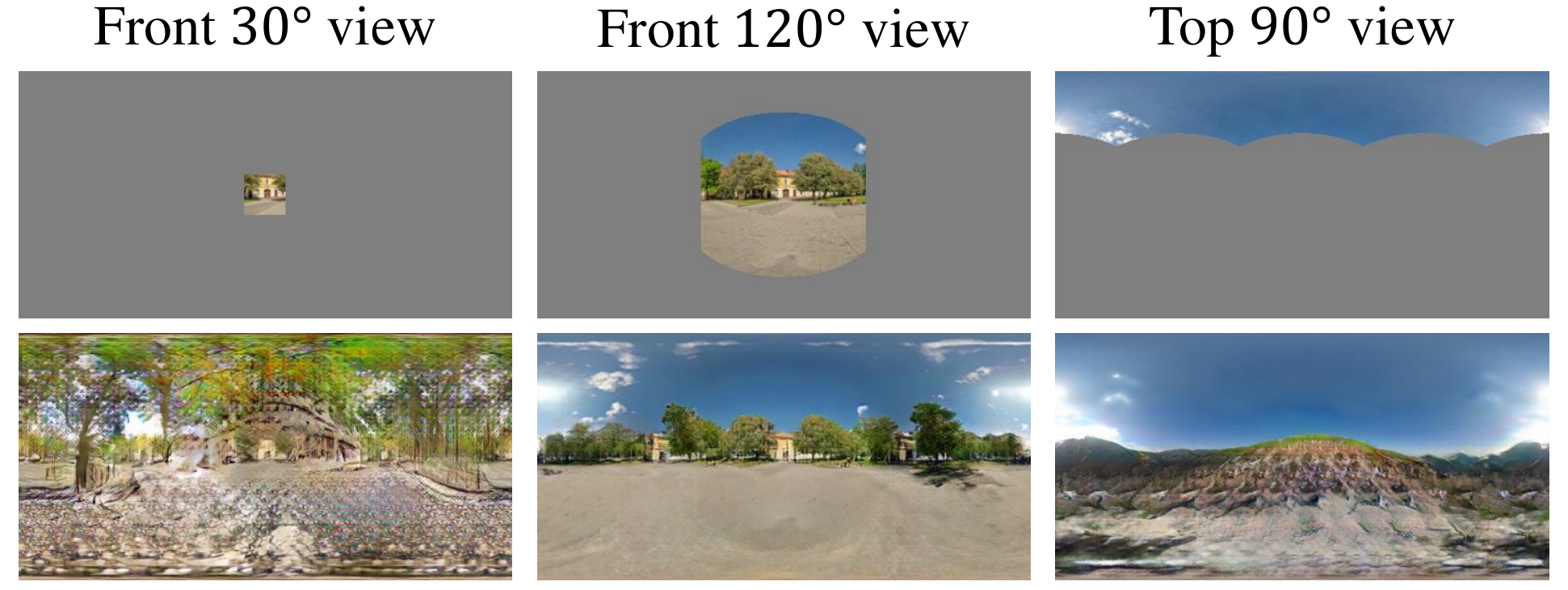}{Generation results from various views for the same ground truth (the fourth image from the left in \refFig{sample}). Top: input images, Bottom: generated images.}{85mm}

\subsubsection{Quantitative evaluation}

\begin{table}[tb]
\caption{Comparison with each method.}
\label{table:quant}
  \centering
  \begin{tabular}{p{3.5em} | p{4.0em} p{4.0em} p{4.0em} } \hline
  \hfil Method & \hfil  Pix2pix & \hfil  PICNet & \hfil  Ours \\ \hline \hline
  \hfil FID & \hfil  46.3 & \hfil  39.6 & \hfil  28.6 \\
  \hline
  \end{tabular}
\end{table}

\begin{table}[tb]
\caption{Comparison with each viewpoint and FOV.}
\label{table:view_fov}
        \begin{center}
            \begin{tabular}{c|cccc|c|c}\hline
             \multicolumn{1}{c|}{Viewpoint} & \multicolumn{4}{c|}{Front} & \multicolumn{1}{c|}{Top} & \multicolumn{1}{c}{Bottom}\\
            \hline
            FOV & 30$^\circ$ & 60$^\circ$ & 90$^\circ$ & 120$^\circ$ & 90$^\circ$ & 90$^\circ$ \\ \hline \hline
            FID & 75.3 & 42.6 & 30.4& 27.4 & 81.7 & 81.5 \\ \hline
        \end{tabular}
    \end{center}
\end{table}

\putFigW{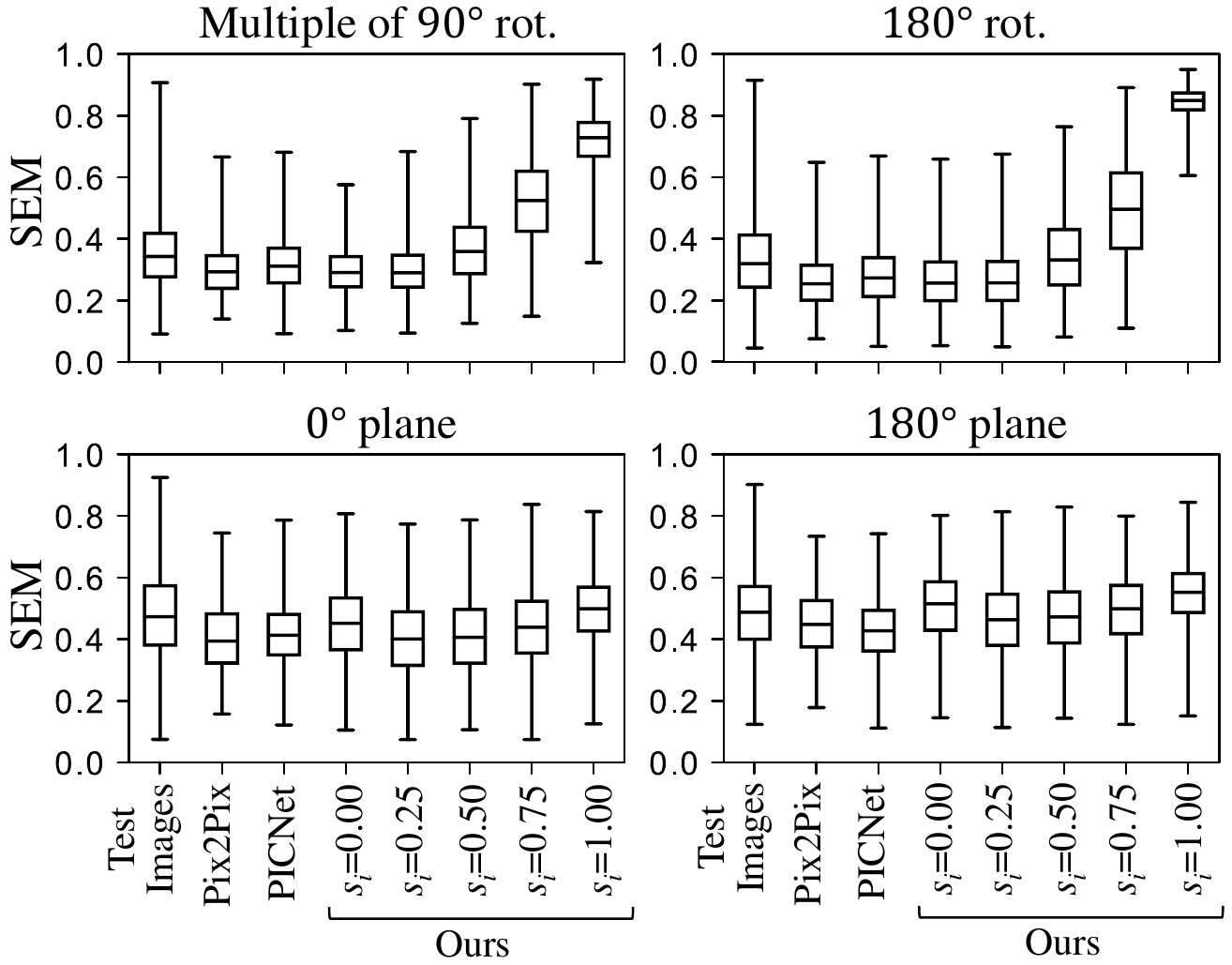}{Symmetry-evaluation metric of the four symmetric types for test images and generated images. The boxplot shows median, quartile, maximum, and minimum.}{80mm}

We evaluated each method by generating 10,000 images from partial images with randomized viewpoints and FOV for the test dataset by using the Frechet inception distance (FID) score \cite{fid}, which measures the distance between the distribution of ground truth and that of the generated images. The symmetry parameter of our proposed method was sampled from the prior in \refEq{p_s}. The evaluation results are presented in Table \ref{table:quant}. Our method displayed a superior FID score than those of the other methods, meaning that our method can generate plausible images by capturing the distribution of high-order features of the spherical images.

We also evaluated the generated image from the fixed view, i.e., the front 30$^\circ$, 60$^\circ$, 90$^\circ$, 120$^\circ$, top  90$^\circ$, and bottom 90$^\circ$ views. The generated samples are depicted in \refFig{view_variation}, and the evaluation metrics are listed in Table \ref{table:view_fov}. We can generate spherical images satisfactorily from a wide FOV. However, the generation results in poor quality from a narrow FOV, and top and bottom viewpoints, as these views do not have sufficient information about the entire image.

We now quantitatively validate the symmetric controlling. Here, we define a symmetry-evaluation metric (SEM) as a normalized autocorrelation between regular features and symmetric-transformed ($T_i$ in Section \ref{sec:symmetry}) features obtained from the fifth block in VGG16 \cite{vgg} for each symmetry type. We set $s_i$ to 0.00, 0.25, 0.50, 0.75, and 1.00 for $i$ corresponding to the targeted symmetric type, and other elements $s_{j \neq i}$ were fixed to low symmetry intensity, i.e., 0.3. Figure \ref{fig:symmetry_eval} depicts the SEM of the four symmetric types for the generated images (SEM for the multiple of 90$^\circ$ rotational symmetry is an average of SEM for 90$^\circ$, 180$^\circ$ and 270$^\circ$ rotations). This proves that varying our symmetry parameter $s$ allows controlling the symmetry of the generated image for a sufficiently wide range, compared with the test images. Furthermore, compared to the rotation symmetry, the control range of the plane symmetry is narrower, as there are some infeasible cases in which the symmetry axis is in the asymmetric partial image (e.g., the eighth row and fifth column image in \refFig{sample}). Furthermore, when the targeted symmetry parameter is 0.00, the SEM is slightly large because it is affected by the other symmetry parameters fixed to 0.3. To validate the plausibility of these generated images, we divided the test images into 2,048 images with higher SEM and 2,048 images with lower SEM, for the four symmetry types, following which we measured the FID scores of the generated images for each dataset, as depicted in \refFig{fid_s}. When $s$ is large at a certain level, it performs well for the high symmetric images.  Furthermore, whichever value is set to $s$, except approximately 0,  our method outperforms the baseline for either high or low symmetric images. Notably, our method outperformed the baseline even for low symmetric images. The circular shift and flip not only control the symmetry but also allow performing convolutions using distant positions of the spherical image (i.e., 180$^\circ$ opposite side), leading to spherical-image generation while considering the consistency of the entire space. 

\putFigW{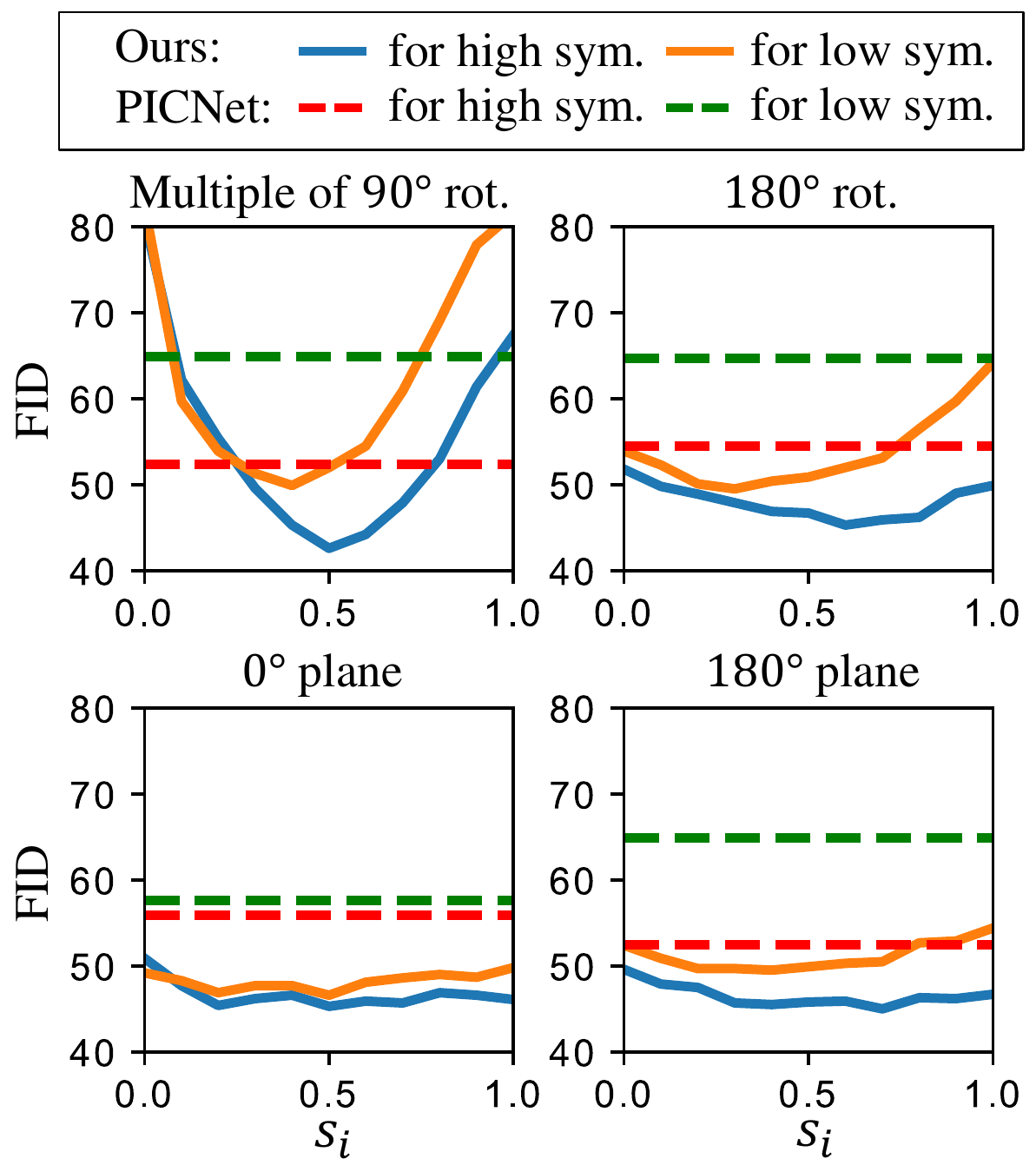}{FID scores for high and low symmetric images.}{60mm}

\subsection{Ablation Study}

\begin{table}[tb]
\caption{Comparison with each loss function. }
\label{table:ablation}
  \centering
  \begin{tabular}{c|ccc} \hline
    Loss & \hfil ${\mathcal L}_{\rm rec}$ and ${\mathcal L}_{\rm gen}$ & \hfil ${\mathcal L}_{\rm rec}$ only & \hfil ${\mathcal L}_{\rm gen}$ only \\ \hline \hline
    L1 & 0.279 & 0.263 & 0.283 \\ \hline
    FID & 28.6 & 41.8  & 31.1 \\ \hline
  \end{tabular}
\end{table}

We compared the generation quality with the evaluation function ${\mathcal L}_{\rm rec}$ in \refEq{app1b}, ${\mathcal L}_{\rm gen}$ in \refEq{app2b} and their combination in \refEq{eval} as shown in Table \ref{table:ablation}. The combination method is superior in terms of FID. The estimated $s$ using ${\mathcal L}_{\rm rec}$ achieved low L1 loss; however, it has less variance. It is conditioned with the input spherical image $x_g$, so the network was not trained for various $s$. On the other hand, using ${\mathcal L}_{\rm gen}$, the network learned with the wide range of $s$ sampled from $p(s)$; however, it did not learn with $s$ corresponding to $x_g$. By combining the two functions, the network can be trained for both the symmetry of $x_g$ and various symmetry with a symmetry parameter. Additional ablation studies are described in the Appendix \ref{sec:ablation_study}.

\section{Conclusion}
We proposed a novel, method to generate spherical-images from a single NFOV image by controlling scene symmetry. We incorporated scene symmetry into CVAEs as a latent variable, and the scene symmetry was implemented as a circular shift of the hidden variables of the neural networks. Furthermore, our experimental results showed that the proposed method can generate various plausible spherical images controlled from symmetric to asymmetric. Future work includes generating high-resolution images and controlling the local features of the generated images.

\subsection*{Acknowledgment}
This work was partially supported by JST CREST Grant Number JPMJCR1403, and partially supported by JSPS KAKENHI Grant Number JP19H01115. We would like to thank Antonio Tejero de Pablos, Atsuhiro Noguchi, Li Yang, Sho Inayoshi, Takuhiro Kaneko, Wataru Kawai and Yusuke Kurose for helpful discussions.

{\small
\bibliographystyle{ieee_initname}
\bibliography{egbib}
}
\appendix
\section{Probabilistic Framework}
\label{sec:prob_framework}

In this section, we describe the details of the probabilistic framework of our method, which generates a spherical image $x_g$ from a partial image (e.g., an NFOV image) $x_l$ by using a scene-symmetry parameter $s \in \mathbb{R}^C$.  We introduce $z \in \mathbb{R}^D$ to a latent variable. \refFig{g_model} shows the assumed graphical model, which indicates the causal relationship between variables. Under this assumption, the joint distribution is expressed as follows:
\begin{eqnarray}
\label{eq:joint}
p(x_g, x_l, z, s) \!\!\!&=&\!\!\! p(x_g | x_l, z, s) p(x_l | z) p(z) p(s) \nonumber \\
                            \!\!\!&=&\!\!\! p(x_g | x_l, z, s) p(z | x_l) p(x_l ) p(s)
\end{eqnarray}
Therefore, the conditional probability for $x_l$ is described as follows:
\begin{eqnarray}
\label{eq:cond}
p(x_g z, s | x_l) \!\!\!&=&\!\!\!  \frac{p(x_g, x_l, z, s)}{p(x_l)}  \nonumber \\
                            \!\!\!&=&\!\!\!  p(x_g | x_l, z, s) p(z | x_l) p(s)
\end{eqnarray}
Next, we introduce the variational lower bound of $\log p(x_g | x_l) $ as follows:
\begin{eqnarray}
\label{eq:elbo}
\log p(x_g | x_l) \!\!\!&=&\!\!\! \log \int q(z, s | x_g, x_l) \frac{p(x_g z, s | x_l)}{q(z, s | x_g, x_l) } dz ds \nonumber \\
                            \!\!\!&\geq&\!\!\! \int q(z, s | x_g, x_l) \log \frac{p(x_g z, s | x_l)}{q(z, s | x_g, x_l) } dz ds \nonumber \\
                            \!\!\!&=&\!\!\! \int q(z, s | x_g, x_l) \nonumber \\
                            \!\!\!& &\!\!\! \times \log \frac{p(x_g | x_l, z, s) p(z | x_l) p(s)}{q(z, s | x_g, x_l) } dz ds \nonumber \\
                            \!\!\!&=&\!\!\! - {\rm KL}(q(z, s | x_g, x_l) || p(z | x_l) p(s)) \nonumber \\
                            &&\!\!\!  + {\rm E}_{q(z, s | x_g, | x_l) } [\log p(x_g | x_l, z, s)]  \nonumber \\
                            \!\!\!&:=& {\mathcal L} 
\end{eqnarray}
where ${\rm KL}(q||p)$ denotes the KL divergence between $q$ and $p$, and ${\rm E}_q[\cdot]$ denotes the expected value for the probability distribution $q$. In the transformational formula, we employed Jensen's inequality for the second row and \refEq{cond} for the third row.

\putFigW{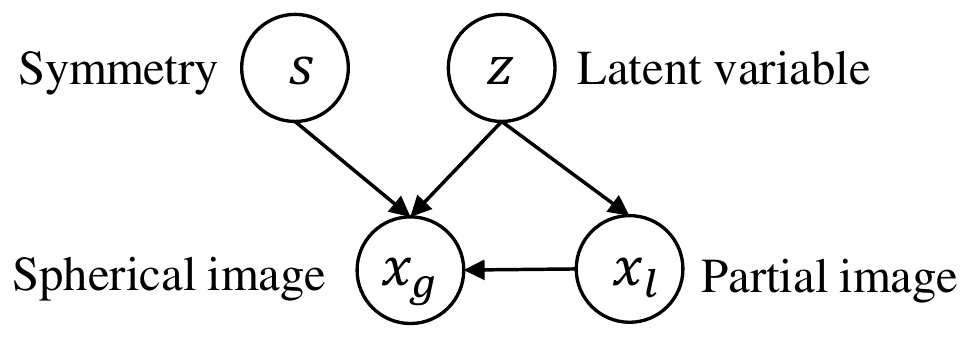}{The graphical model indicates the causal relationship of a spherical image, a partial image, a latent variable, and a symmetry parameter.}{60mm}

To maximize the variational lower bound, we employed the following two types of decomposition approximations. This is because it is difficult to perform the maximization directly. We have
\begin{eqnarray}
\label{eq:app1_ap}
q(z, s|x_g, x_l) &=& q(z|x_g) q(s|x_g) \\
\label{eq:app2_ap}
q(z, s|x_g, x_l) &=& q(z|x_l) q(s)
\end{eqnarray}
Our method can generate spherical images using the approximation in \refEq{app1}. The right-hand side of \refEq{elbo} can be transformed as follows:
\begin{eqnarray}
\label{eq:app1b_ap}
{\mathcal L_1} \!\!\!\!&=&\!\!\!\! -{\rm KL}(q(z |x_g) || p(z|x_l)) \nonumber \\
&&\!\!\!\!-{\rm KL}(q(s |x_g) || p(s))  \nonumber \\
&&\!\!\!\!+ {\rm E}_{q(z|x_g) q_(s|x_g)} [p(x_g | x_l, z, s)]
\end{eqnarray}
On the other hand, using the approximation in \refEq{app2_ap}, the right-hand side of \refEq{elbo} can be transformed as follows:
\begin{eqnarray}
\label{eq:app2b_ap}
{\mathcal L_2} \!\!\!\!&=&\!\!\!\! -{\rm KL}(q(z |x_l) || p(z|x_l)) \nonumber \\
&&\!\!\!\!-{\rm KL}(q(s) || p(s))  \nonumber \\
&&\!\!\!\!+ {\rm E}_{q(z|x_l) q(s)} [p(x_g | x_l, z, s)]
\end{eqnarray}
${\rm KL}(q || p )$ takes the minimum value of 0 when $q = p$. Furthermore, \refEq{app2b_ap} can be transformed as follows:
\begin{eqnarray}
\label{eq:app2c_ap}
{\mathcal L_2}= {\rm E}_{p(z|x_l)p(s)} [p(x_g | x_l, z, s)]
\end{eqnarray}
For the expression of $p (x_g | x_l, z, s)$, we employ $l_\omega^{\rm rec}(x_g, x_l, z, s)$ and $l_{\omega}^{\rm gen}(x_l, z, s)$ for ${\mathcal L_1}$ and ${\mathcal L_2}$ respectively. Thereafter, we get the reconstruction likelihood ${\mathcal L}_{\rm rec}$ and the generation likelihood ${\mathcal L}_{\rm gen}$ in the body of this paper. 

\section{Generated Samples}
\label{sec:additional_examples}
\subsection{Additional Examples}

We show the additional examples of generated spherical images in \refFig{sample1}, \refFig{sample2} and \refFig{sample3}. The viewpoints of the input image are randomized uniformly in the sphere, and FOV is randomized uniformly from 30$^\circ$ to 120$^\circ$. In our method, the symmetry parameter $s \in \mathbb{R}^5$ was set to five variations, namely, $(h, h, h, l, l)$, $(l, h, l, l, l)$, $(l, l, l, h, l)$, $(l, l, l, l, h)$, and $(l, l, l, l, l)$, where $h=1.0, l=0.3$, for the multiple of 90$^\circ$ and 180$^\circ$ rotational symmetries, the plane symmetries to the 0$^\circ$ and 90$^\circ$ axes, and asymmetry, respectively . As shown in these figures, our method can generate spherical images with various symmetric types from a single NFOV image, even if the ground truth is asymmetry. Although it is difficult to reconstruct ground truth especially from an image with a viewpoint near pole or a narrow FOV, plausible spherical images are generated in most cases. Our method is useful for generating the plausible spherical image that has the desired specific symmetric type including asymmetry.

\subsection{Transformation from Symmetry to Asymmetry}

We show the samples that are controlled continuously from symmetric to asymmetric in \refFig{trans}. We set the $s_i$ to 0.00, 0.25, 0.50, 0.75 and 1.00 and $s_{j \neq i} = 0.3$, where $i$ is the index of the targeted symmetry type.  It indicates that our method can change the symmetry continuously. When $s_i = 0.00$ for 90$^\circ$ rotational symmetry, the quality of generated image is low. These results are corresponding to FID score in Fig. 11 in the body. When $s_i = 0.00$ for 0$^\circ$ plane symmetry or 90$^\circ$ plane symmetry, there are influences of other types of symmetry as $s_{j \neq i} = 0.3$. These results are corresponding to higher SIM than $s_i = 0.25$ in Fig. 10 in the body, because other types of symmetry have a little effect on SEM.

\section{Ablation Study}
\label{sec:ablation_study}

\subsection{Circular Padding}
\putFigW{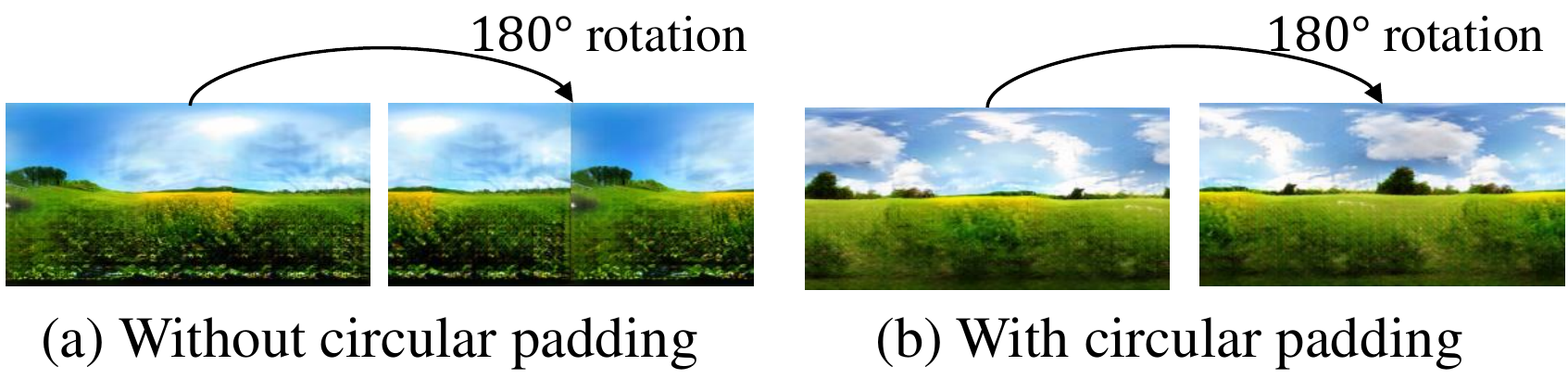}{Effect of the circular padding}{82mm}
Figure \ref{fig:sample_cp} indicates the difference in generated images depending on the presence or absence of the circular padding. The generated image is rotated 180$^\circ$ around the gravity axis so that the joined portion can be seen, and the left and right ends are displayed in the center. From the figure, it can be seen that the discontinuity at the left and right ends is eliminated by the circular padding.

\subsection{Symmetry Control}
We compare the generation quality with and without the symmetry control function. We note that we train the network from scratch without the symmetry control function, which is different from setting $s=0$ in the learned network with the symmetry control function. In addition, we compare the networks learned with and without the weight ($w_{\kappa}$ in Eq. (16)) corresponding to the partial image region in the symmetry control function.  Table \ref{table:ablation_sym} shows the FID for each setting. It reveals that the symmetry control function is effective for improving plausibility because it can generate symmetric images in test data and allows performing convolutions using distant positions of the spherical image even if the symmetry intensity is low. In addition, the weight corresponding to the partial image region is useful because it can use information in highly reliable areas with emphasis.

\begin{table}[tb]
\caption{Comparison with the symmetry control function. }
\label{table:ablation_sym}
  \centering
  \begin{tabular}{c|ccc} \hline
    & Ours & w/o sym. control &  w/o weight \\ \hline \hline
    FID & 28.6 & 40.3 & 37.2 \\ \hline
  \end{tabular}
\end{table}

\putFigWWH{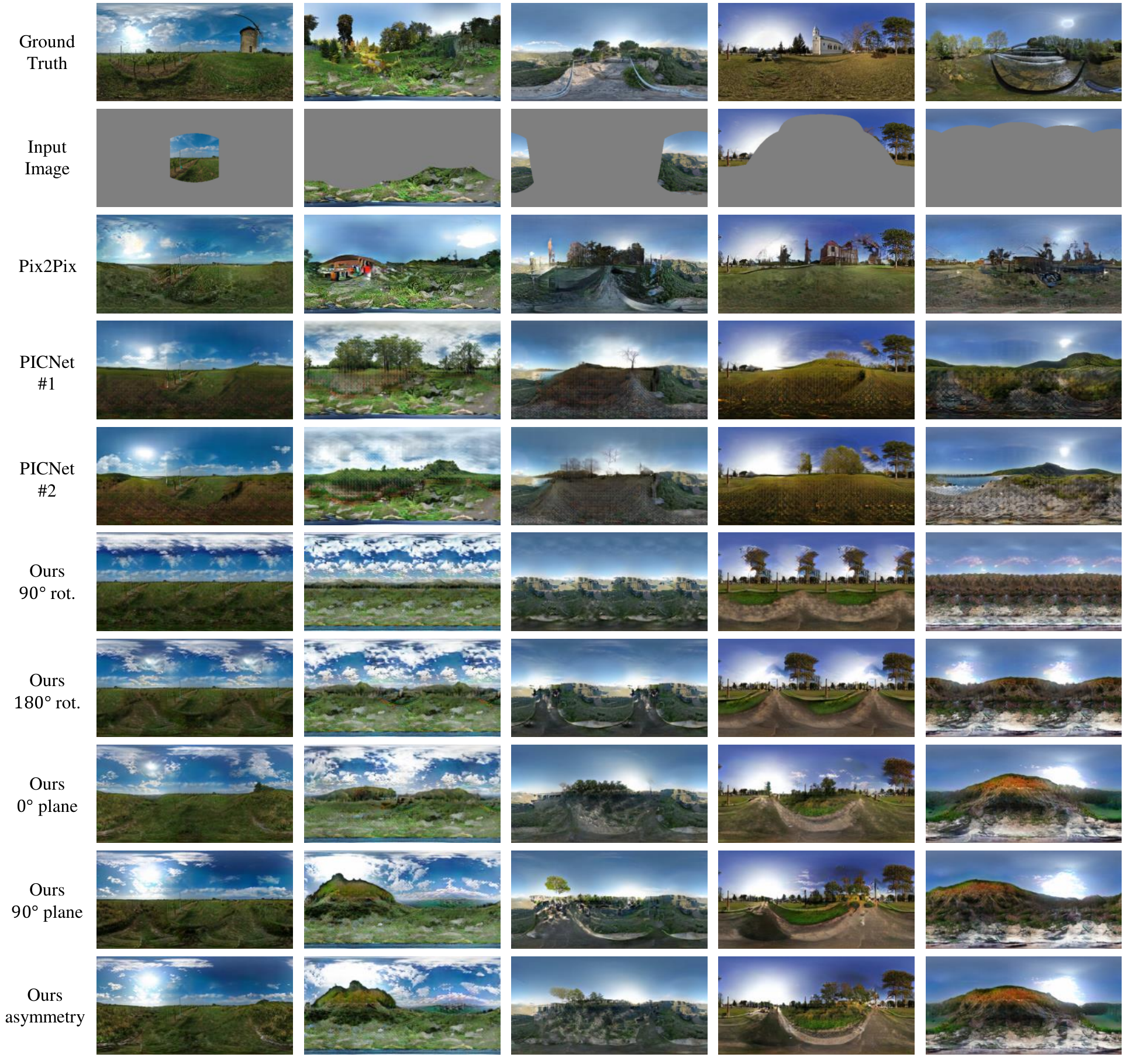}{Additional examples of spherical images generated using our method and previous works for natural scene.}{175mm}
\putFigWWH{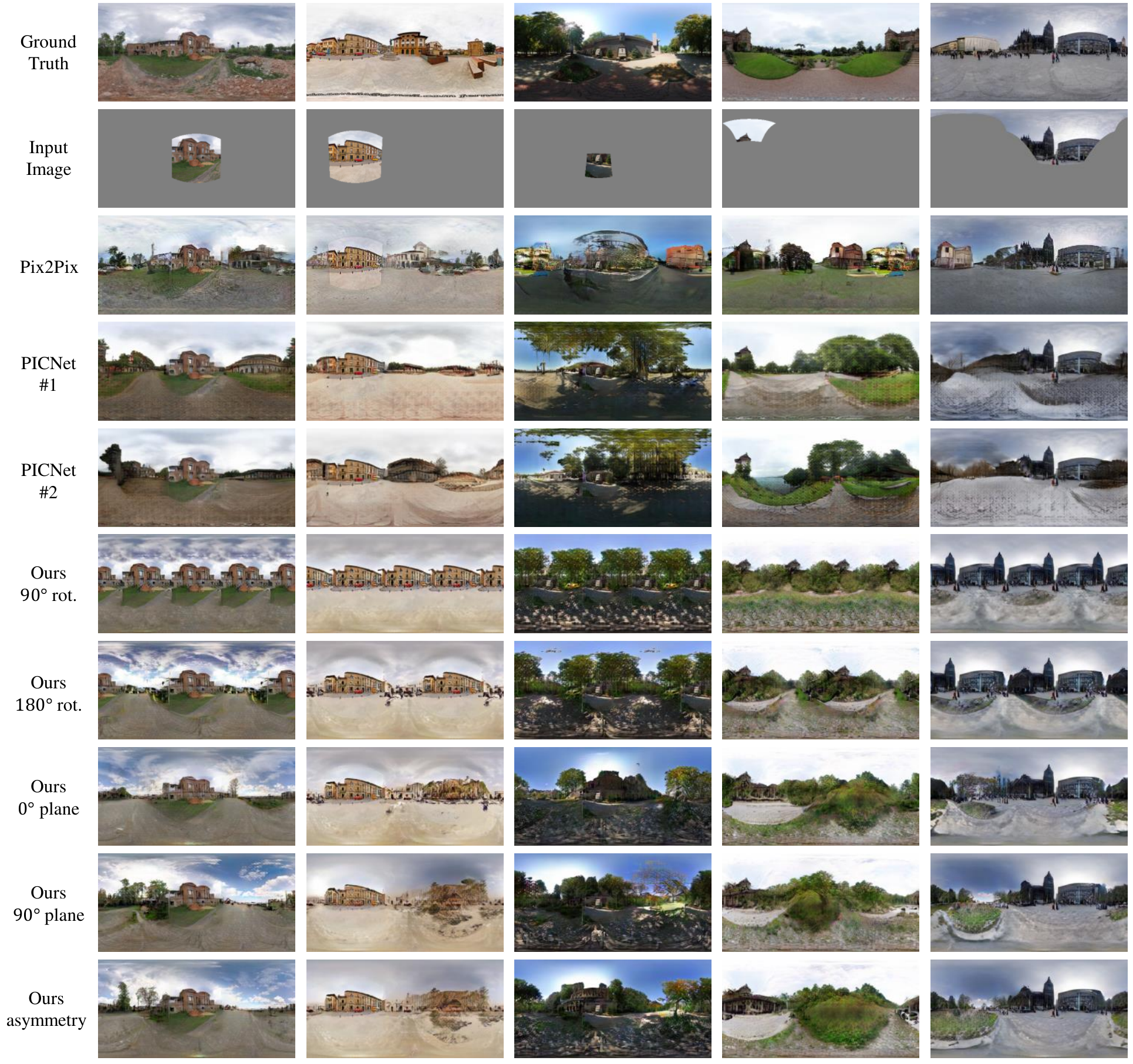}{Additional examples of spherical images generated using our method and previous works for the scene including buildings}{175mm}
\putFigWWH{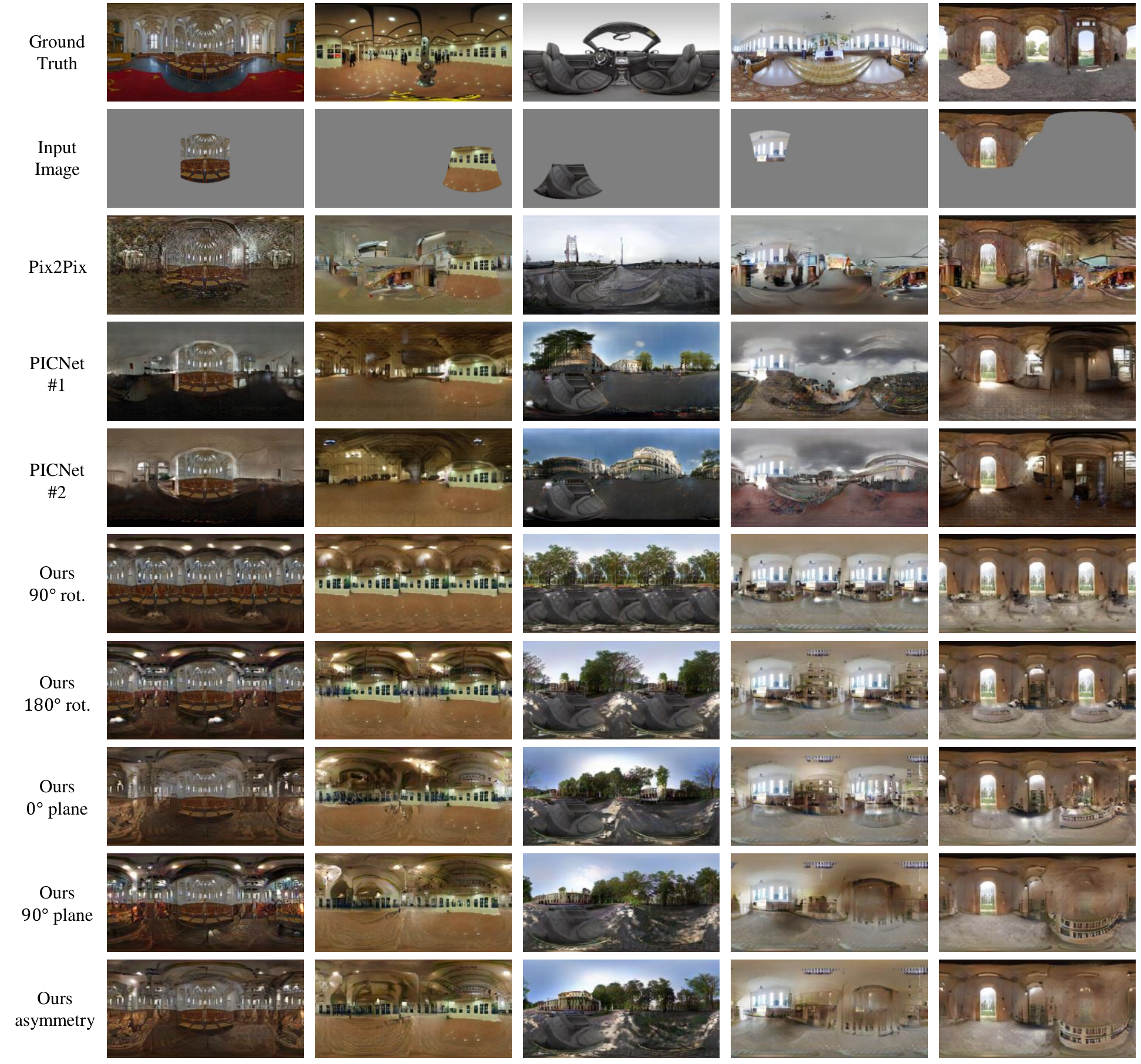}{Additional examples of spherical images generated using our method and previous works for indoor scene.}{175mm}
\putFigWWH{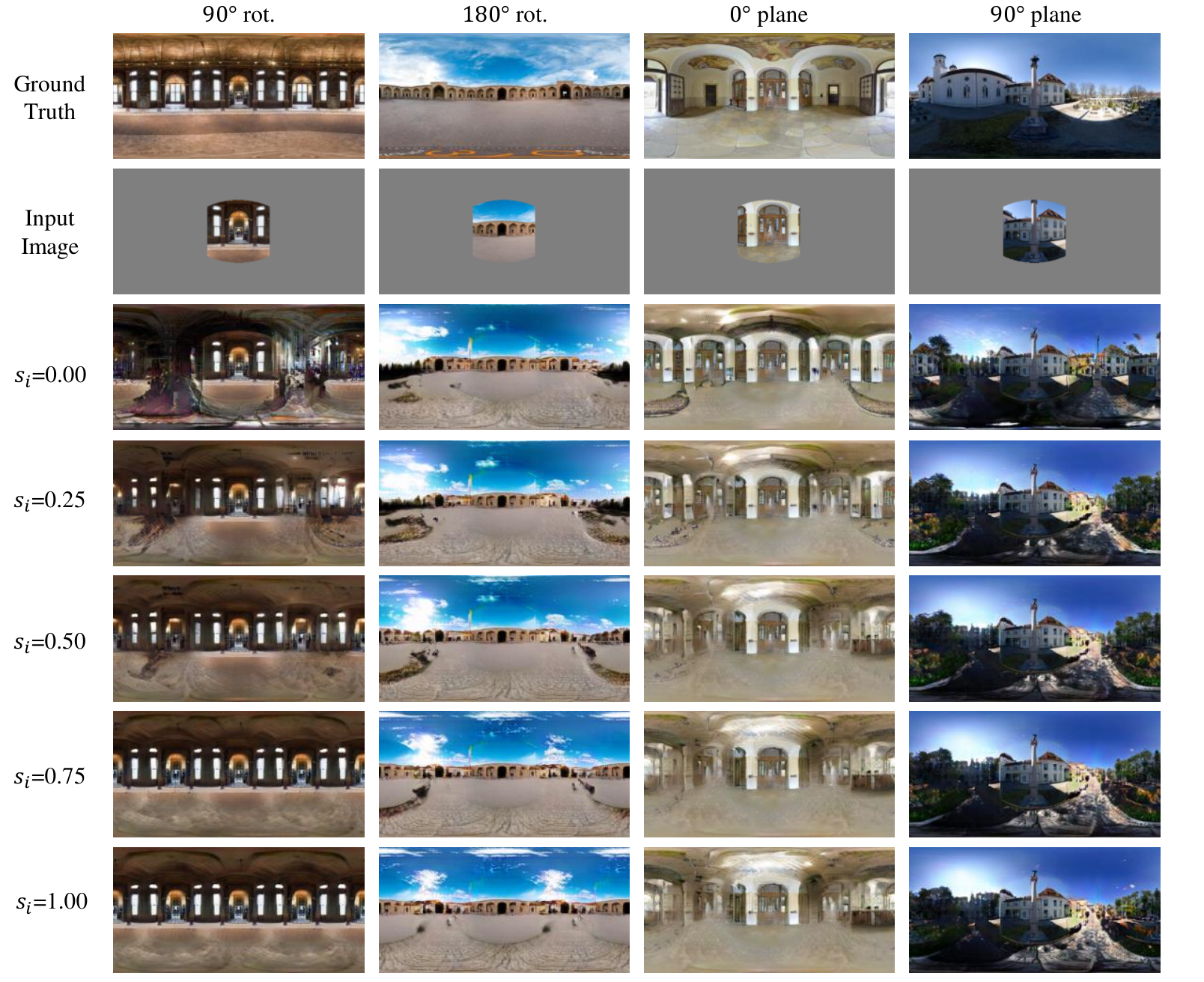}{The generated spherical images that are controlled continuously  from symmetric to asymmetric.}{160mm}

\end{document}